\documentclass{article}

\usepackage{arxiv}
\usepackage{wrapfig}
\usepackage[utf8]{inputenc} % allow utf-8 input
\usepackage[T1]{fontenc}    % use 8-bit T1 fonts
\usepackage{hyperref}       % hyperlinks
\usepackage{url}            % simple URL typesetting
\usepackage{booktabs}       % professional-quality tables
\usepackage{amsfonts}       % blackboard math symbols
\usepackage{nicefrac}       % compact symbols for 1/2, etc.
\usepackage{microtype}      % microtypography
\usepackage{lipsum}		% Can be removed after putting your text content
\usepackage{graphicx}
\usepackage{natbib}
\usepackage{doi}
\usepackage{float}
\usepackage{caption}
\usepackage{subcaption}
\usepackage{booktabs,chemformula}
\usepackage{multirow}

\usepackage[utf8]{inputenc}
\usepackage[T1]{fontenc}
\usepackage{multirow}
\usepackage{colortbl}
\usepackage{comment}
\usepackage{bm}
\usepackage{amsmath}

\title{MAD-CNN: High-Sensitivity and Robust Collision Detection for Robots with Variable Stiffness Actuation}

\author{$^1$Zhenwei Niu, $^2$Lyes Saad Saoud and ~$^{3}$Irfan Hussain, \\
	Khalifa University Center for Autonomous and Robotic Systems, Khalifa University, \\
 Advanced Research and Innovation Center, Khalifa University, \\Abu Dhabi, United Arab Emirates, P O Box 127788, Abu Dhabi, UAE \\
        \texttt{$^1$zhenwei.nui@ku.ac.ae, $^2$lyes.saoud@ku.ac.ae, $^{3}$irfan.hussain@ku.ac.ae}
}

\begin{document}
\maketitle
\begin{abstract}
Safe and efficient human-robot collaboration (HRC) relies on proactive collision avoidance and robust detection of unexpected contacts. Although existing learning-based approaches excel in specific areas, they often encounter challenges with robots equipped with variable stiffness actuators—a crucial aspect for ensuring safe and adaptable HRC. To address this, we present MAD-CNN, a novel Modularized Attention-Dilated Convolutional Neural Network designed for highly sensitive and robust collision detection in these adaptable robots. MAD-CNN leverages a dual inductive bias mechanism, combining modularization and dilated convolution, to significantly reduce data requirements. This capability facilitates efficient training even with limited collision datasets, offering a substantial advantage for real-world implementation. Additionally, an attention module prioritizes relevant features, enhancing the network's robustness against variations in stiffness levels—a core challenge in variable stiffness robots.
This powerful combination unlocks exceptional performance for MAD-CNN. In extensive evaluations, it achieved zero missed collisions across 516 events, with a minimal detection delay averaging just 12.05 milliseconds and 20\% fewer false positives compared to baseline methods. Remarkably, this outstanding performance was attained even when trained on a four-minute dataset focused on the highest stiffness setting, underscoring MAD-CNN's data efficiency and adaptability across diverse stiffness configurations. By enabling accurate and sensitive collision detection across the spectrum of variable stiffness conditions, MAD-CNN paves the way for safer and more reliable HRC in various fields, including manufacturing, healthcare, and assistive robotics.
\end{abstract}


\section {Introduction}

The rapidly growing collaboration between humans and robots in diverse industrial and daily activities has driven a significant surge in physical human-robot interaction (pHRI) tasks. Ensuring safety in these collaborative endeavors is paramount for the effective integration and widespread adoption of human-robot collaboration \cite{zacharaki2020safety}. The focal point of this safety assurance lies in addressing unexpected collisions during such physical interactions.

While the integration of external sensors has proven effective in mitigating unexpected collisions \cite{flacco2012depth, khansari2012dynamical, xu2021motion}, the inherent challenges in predicting and responding to rapid and unpredictable relative motions between humans and robots necessitate a comprehensive approach. In addition to collision avoidance techniques, rapid and accurate collision detection and response mechanisms become indispensable components in the pursuit of maintaining safety during pHRI \cite{haddadin2017robot}.
\begin{figure*}[t]
    \centering
    \includegraphics[width=\linewidth]{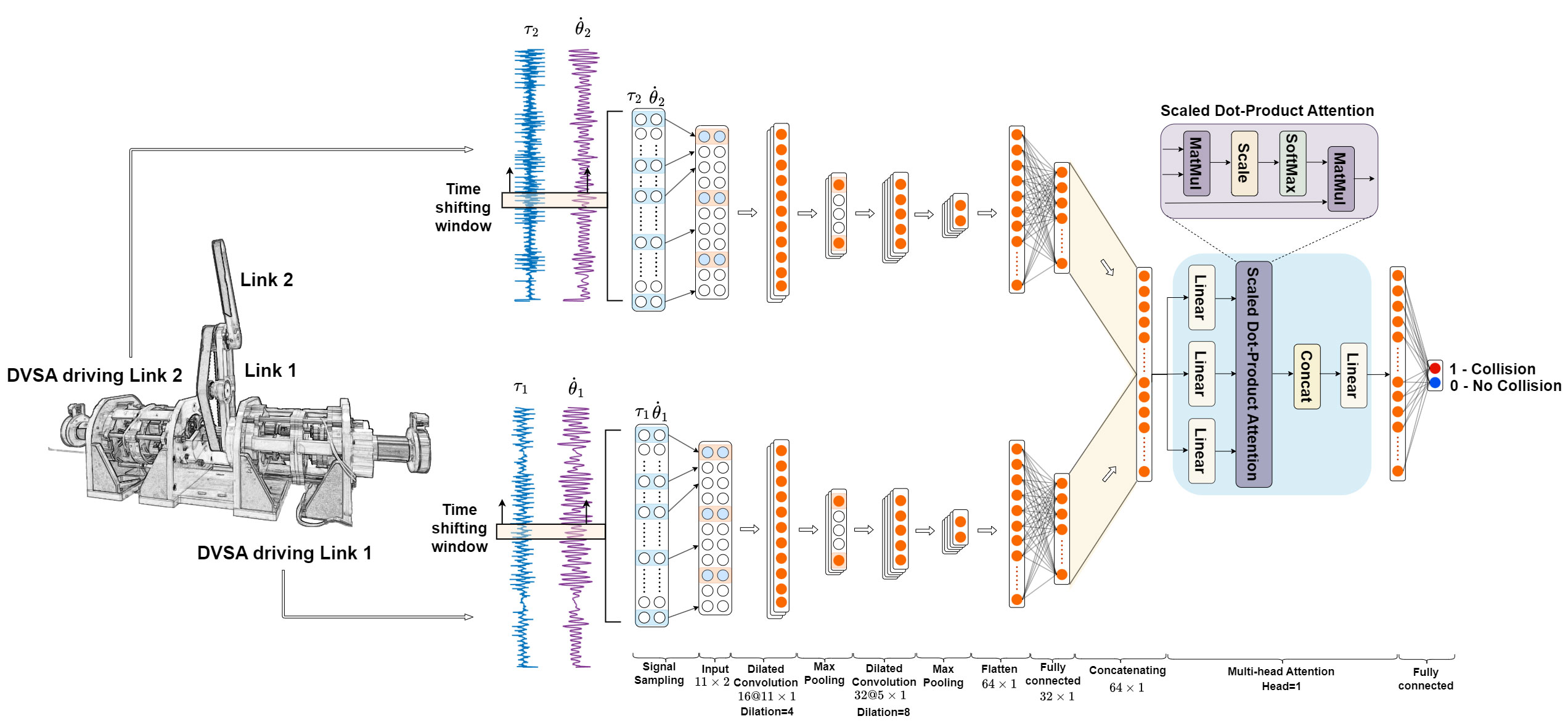}
    \caption{The proposed MAD-CNN architecture for robot collision detection encompasses a comprehensive structure aimed at improving collision sensitivity, robustness, and data efficiency for robots equipped with variable stiffness actuation. MAD-CNN integrates a dual inductive bias mechanism that incorporates joint modularization and a dilated convolutional neural network, along with an attention module. It is well suited for scenarios involving hard-to-collect and imbalanced collision data.}
    \label{fig:network structure}
\end{figure*}
In this context, current collision detection techniques can be broadly categorized into three main groups: artificial skin-based, model-based, and data-driven methods. Artificial skins, constructed from deformable materials and tactile sensors, have demonstrated accurate collision detection capabilities \cite{skin_duchaine2009flexible, skin_pugach2016touch, skin_vergara2018incorporating, skin_teyssier2021human}. However, the challenges associated with the fabrication and coverage of entire robots with these skins pose obstacles in terms of complexity and cost.

Conversely, model-based approaches, exemplified by the generalized momentum observer (MOB) method, leverage an identified dynamic model to estimate external torque exerted on robots, determining collision presence based on a user-defined threshold \cite{de2003actuator, de2006collision, birjandi2020observer}. Despite the attention garnered by MOB, model-based techniques necessitate accurate identification of the dynamic robot model, which is hindered by the nonlinear nature of friction and the presence of unmodeled effects or uncertainties like backlash and elasticity \cite{park2020learning, kim2021transferable}. Adjusting the user-defined threshold also demands considerable effort.

In response to these challenges, learning-based collision detection techniques have emerged as alternatives to traditional model-based methods. By training neural network models with collision and collision-free data, these techniques accommodate uncertainties and unmodeled effects within the dynamic model, eliminating the need for a user-defined threshold \cite{park2020learning}. Learning-based methods are particularly prevalent in commercial robots with rigid joints, offering promising results in terms of collision detection accuracy.

However, a critical gap exists in the validation and adaptation of these learning-based methods for collaborative robots equipped with variable stiffness actuation. Collaborative robots (cobots) with variable stiffness actuation have the potential for more intimate human collaboration, heightening the urgency to address the challenges associated with accurate, rapid, and robust collision detection in this context. The dynamic changes in stiffness during tasks pose additional difficulties for collision detection, making traditional learning-based approaches less suitable for these scenarios.

Moreover, collecting sufficient collision data for training purposes poses challenges and potential hazards for both human users and robots. Therefore, there is a pressing need for data-efficient solutions that can navigate these challenges and optimize classification accuracy, especially in the face of imbalanced datasets.

In light of these challenges, this paper introduces MAD-CNN, a novel network architecture designed for collision detection in robots equipped with variable stiffness actuators. MAD-CNN incorporates a dual inductive bias, comprising joint modularization and dilated convolutional neural network, along with an attention module, to address the intricacies associated with variable stiffness actuation. The proposed architecture aims to enhance collision sensitivity, robustness, and data efficiency, particularly in scenarios involving hard-to-collect and imbalanced collision data.

\subsection*{Related Works}

Previous research in the realm of learning-based collision detection has predominantly focused on commercial robots with rigid joints, demonstrating efficacy in residual and uncertainty regression or implementing end-to-end approaches \cite{sharkawy2018manipulator, czubenko2021simple, xu2020new, lim2021momentum}. While these methods have proven effective for their intended applications, their direct transferability to collaborative robots with variable stiffness actuation remains uncertain.

The challenges associated with collecting diverse and sufficient collision data for training purposes, coupled with the imbalanced nature of the collected datasets, necessitate novel approaches to enhance data efficiency. Modularized neural networks (MNN), as introduced in \cite{kim2021transferable}, have shown promise in managing imbalanced datasets but still require substantial amounts of collision and free-motion data for training. This underscores the need for more efficient and data-effective solutions in the context of collision detection.

In response to these challenges, MAD-CNN presents a unique contribution through the incorporation of joint modularization, dilated convolution, and an attention module to create a comprehensive collision detection architecture. By addressing the specific challenges posed by variable stiffness actuators, MAD-CNN seeks to overcome the limitations of existing methods and offer a robust solution for collision detection in collaborative robots.

In the subsequent sections of this paper, we delve into the details of MAD-CNN's architecture, its experimental validation, and its performance compared to existing methods, highlighting its potential contributions to the field of human-robot collaboration and safety in pHRI tasks.

\subsection*{Main Contributions and Organization of the Paper}
This paper presents the Modularized Attention-Dilated Convolutional Neural Network (MAD-CNN), a novel collision detection algorithm for robots with variable stiffness actuators (VSAs). MAD-CNN leverages a dual inductive bias mechanism and an attention module to achieve remarkable performance with minimal training data and enhanced robustness across diverse stiffness conditions. Our key contributions are as follows:
\begin{itemize}
     \item \textbf{Highly efficient collision detection:} MAD-CNN achieves 100\% collision detection accuracy and minimal detection delay across all stiffness conditions while trained on data from only a single stiffness setting. This signifies exceptional data efficiency, requiring only 4 minutes of collision data for training.
     \item \textbf{Robustness to imbalanced data:} MAD-CNN effectively tackles highly imbalanced datasets by successfully detecting all 516 collisions (30 minutes of data) occurring at random links during various robot motions and stiffness levels.
     \item \textbf{Superior performance:} Compared to state-of-the-art methods with the same limited training data, MAD-CNN exhibits enhanced collision sensitivity and robustness, effectively minimizing the common issue of false positives in learning-based approaches. An ablation study further validates the efficacy of MAD-CNN's structure, solidifying its superior performance.
\end{itemize}

The rest of this paper is organized as follows: Section \ref{sec:platform introduction} provides an introduction to the platform used in this study. The proposed MAD-CNN model is elaborated in detail in Section \ref{sec:MAD-CNN structure}. Data collection and input processing are described in Section \ref{Section: data collection}. In Section \ref{sec: results}, we present the comprehensive evaluation of MAD-CNN, including the ablation study and a comparison with state-of-the-art models. Additionally, the experimental results showcasing the integration of MAD-CNN with the continuous filter (CF) validate the collision detection sensitivity and robustness of our proposed model. Finally, Section \ref{sec:conclusion} concludes this study.  

\section*{Manipulator with Variable Stiffness Actuation}
\label{sec:platform introduction}

\begin{figure}[t]
    \centering
    \includegraphics[width=\linewidth]{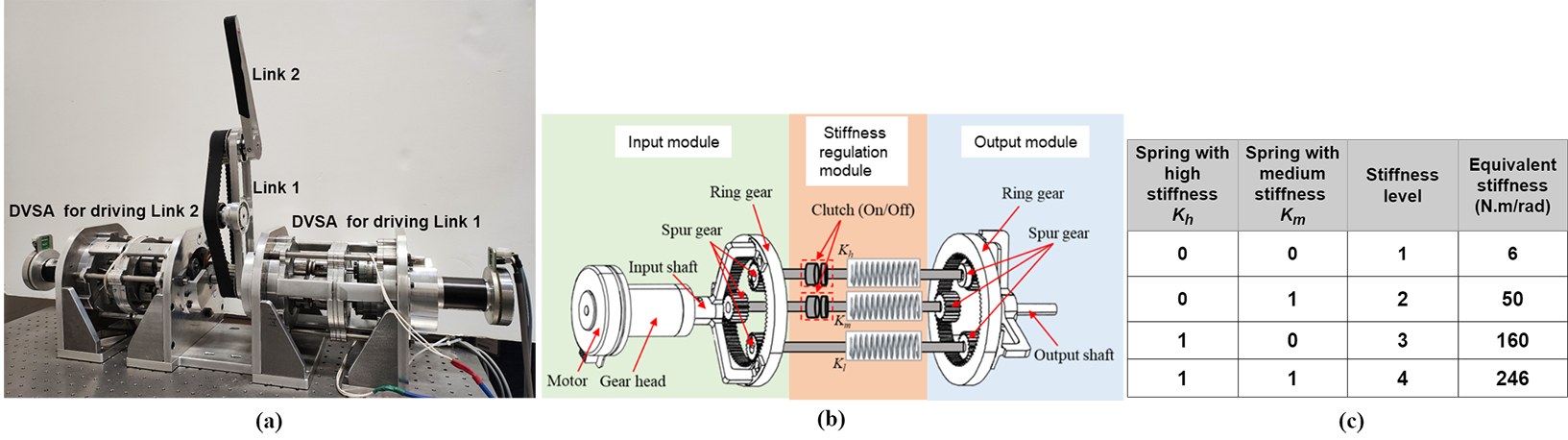}
    \caption{Experiments platform. (a) A robotic manipulator with discrete variable stiffness actuators (DVSAs). (b) The working principle of the DVSA. It has four levels of intrinsic physical stiffness, and the stiffness can be rapidly adjusted online by controlling the engagement of the springs. (c) Joint Stiffness Levels of DVSA}
    \label{fig:manipulator and DVSA working principle}
\end{figure}

This study employs a robotic manipulator equipped with discrete variable stiffness actuators (DVSAs). The detailed design can be found in our previous work \cite{niu2022towards}, in which we introduced an innovative collision-handling pipeline. The pipeline incorporates a DVSA-based manipulator to effectively mitigate the impact of unexpected collisions and improve safety during human-robot interaction. In this section, we recall the main design concept. The overall platform configuration and the working principle of the DVSA are shown in Figure \ref{fig:manipulator and DVSA working principle}. Each robot joint has four different levels of physical stiffness. By selectively engaging or disengaging the springs, the stiffness of each joint can be rapidly adjusted to any predefined level. The table in Figure \ref{fig:manipulator and DVSA working principle} (c) presents a comprehensive overview of the specific values associated with joint stiffness levels. Notably, in our previous study \cite{niu2022towards}, one of the important reaction strategies involves rapidly switching the physical stiffness of the joint 
from its current level to the lowest level (stiffness level 1, as shown in Figure \ref{fig:manipulator and DVSA working principle} (c)). Experimental validation was conducted to evaluate the effectiveness of this strategy in enhancing safety. This study focuses on collision detection for collaborative robots equipped with variable stiffness actuation, and our model is proposed based on the aforementioned platform to examine the effectiveness of collision detection in this context.   

%%%%%%%%%%%%%%%%%%%%%%%%%%%%%%%%%%%%%%%%%%%%%%%%%%%%%%%%%%%%%%%%%%%%%%%%%%%%%%%%%%%%%%%%%%%%
\section{The Proposed MAD-CNN Model}
\label{sec:MAD-CNN structure}
The proposed network structure, MAD-CNN, for robot collision detection can be found in Figure \ref{fig:network structure}. MAD-CNN is designed to enhance data training efficiency and collision detection robustness for collaborative robots equipped with variable stiffness actuation. Efficient utilization of collision data is crucial due to the costly and laborious nature of its collection process. Moreover, the highly imbalanced features observed in the collected dataset, where the number of collision cases is substantially lower than that of collision-free cases, further emphasize the importance of optimizing data efficiency. Therefore, it is advisable to employ efficient classification methods to accurately distinguish between collision and collision-free cases, even when limited collision information is available \cite{kim2021transferable}. Moreover, collecting collision data for all stiffness conditions is extremely challenging and expensive when dealing with robots equipped with variable stiffness actuators. There is a need to develop a network architecture that is robust to changes in the physical stiffness of the joint.
%%%%%%%%%%%%%%%%%%%%%%%%%%%%%%%%%%%%%%%%%%%%%%%%%%%%%%%%%%%%%%%%%%%%%%%%%%%%%%%%%%%%%%%%%%%%
\subsection{Network Structure}
MAD-CNN employs dual inductive bias in the network structure. The first dual inductive bias, which is the modularized joint network structure, enables the extraction of collision information for each joint through the utilization of local joint variables. This bias in the network structure achieves the decoupling of individual joints by independently inputting joint-specific variables into their corresponding modular networks, which can decrease the search space for network parameters \cite{kim2021transferable}. 

An additional inductive bias used in this study is the utilization of dilated convolutional neural networks. Dilated convolution, as described in \cite{Yu_2017_CVPR}, is a variant of the convolutional layer that offers an expanded receptive field while maintaining the same number of parameters. For a given input $F(s)$, the dilated convolution can be defined as:
\begin{equation}
    (F \ast_{l} k)(p) = \sum_{s+lt=p} F(s)k(t),
\end{equation}
where $\ast_{l}$ is an $l$-dilated convolution, $k(t)$ is the applied filter. With dilated convolution, the network becomes capable of capturing long-range dependencies within the input collision and noncollision signals without sacrificing resolution or computational efficiency. This emphasis on discriminating between these two classes enhances the network's sensitivity to detecting collisions, thereby further improving its collision detection capabilities. The effects of both inductive bias are further validated in Section \ref{sec:ablation study}.

Following the application of modularized dilated joint networks, the resulting extracted features are combined through concatenation and subsequently inputted into a multi-head self-attention module. The self-attention module represents a critical architectural component within the Transformer framework, as highlighted in \cite{NIPS2017_3f5ee243}. In the context of collision detection, the utilization of the self-attention module allows for the prioritization of the most pertinent components within the concatenated joint collision features. This prioritization aids in the classification process by assigning higher weights to the relevant parts of the collision features. For a given input $I \in \mathbb{R}^{d \times n}$, where $d$ represents the input dimension and $n$ denotes the input length, the self-attention mechanism begins by performing a linear projection to derive the components of query ($Q$), key ($K$) and value ($V$), as outlined below:

\begin{equation}
    Q = W_qI, \quad K = W_kI, \quad \text{and} \quad V = W_VI,
\end{equation}
where $W_q$ and $W_k \in \mathbb{R}^{s1 \times d}$, $W_V \in \mathbb{R}^{s \times d}$ are learnable weight matrices. Then the scaled Dot-product is processed to get the attention score:

\begin{equation}
    Z = \text{softmax}(\frac{QK^T}{\sqrt{d_k}})V,
\end{equation}
where $d_k$ corresponds to both the query and key dimensions and it serves the purpose of mitigating the issue of vanishing gradients that may arise when applying the softmax function. After that, by concatenating multi-head information and doing the linear projection, several levels of correlation information can be extracted and used for the final classification.
%%%%%%%%%%%%%%%%%%%%%%%%%%%%%%%%%%%%%%%%%%%%%%%%%%%%%%%%%%%%%%%%%%%%%%%%%%%%%%%%%%%%%%%%%%%%
\subsection{Network Structure Details}
The modularized joint network consists of two 1D dilated convolutional layers. The first layer employs 16 filters, while the second layer employs 32 filters. Both convolutional layers have a filter size of $3$ and a stride of 1. The dilation factor is set to 4 for the first convolutional layer and to 8 for the second convolutional layer. Additionally, max pooling layers are applied after each convolutional layer. The outputs from the max-pooling layers are subsequently flattened and connected to a fully connected layer with 32 hidden neurons. The activation function employed for the fully connected layer is the Gaussian error linear unit (GELU) function, which has been shown to achieve higher accuracy compared to the rectified linear unit (ReLU) function, as demonstrated in \cite{hendrycks2023gaussian}. The resulting outputs from this layer, representing each joint, are then concatenated to form a combined representation comprising 64 hidden neurons. This concatenated representation is then passed through a self-attention module, which features 1 head. The output of the self-attention module is connected to another fully connected layer with 64 hidden neurons, employing the GELU activation function. Finally, the output layer utilizes the softmax function to provide prediction scores between 0 and 1, facilitating the classification of either the "no collision" or "collision" class. The proposed network structure requires only 0.046 $\mu s$ to be computed on an Intel i9-19700H with Nvidia GeForce RTX 3070 Ti, enabling real-time collision detection.
%%%%%%%%%%%%%%%%%%%%%%%%%%%%%%%%%%%%%%%%%%%%%%%%%%%%%%%%%%%%%%%%%%%%%%%%%%%%%%%%%%%%%%%%%%%%
\section{Data Collection and Network Training}
\label{Section: data collection}
%\subsection{Data Collection}
\begin{figure}[t]
    \centering
    \includegraphics[width=0.85\linewidth]{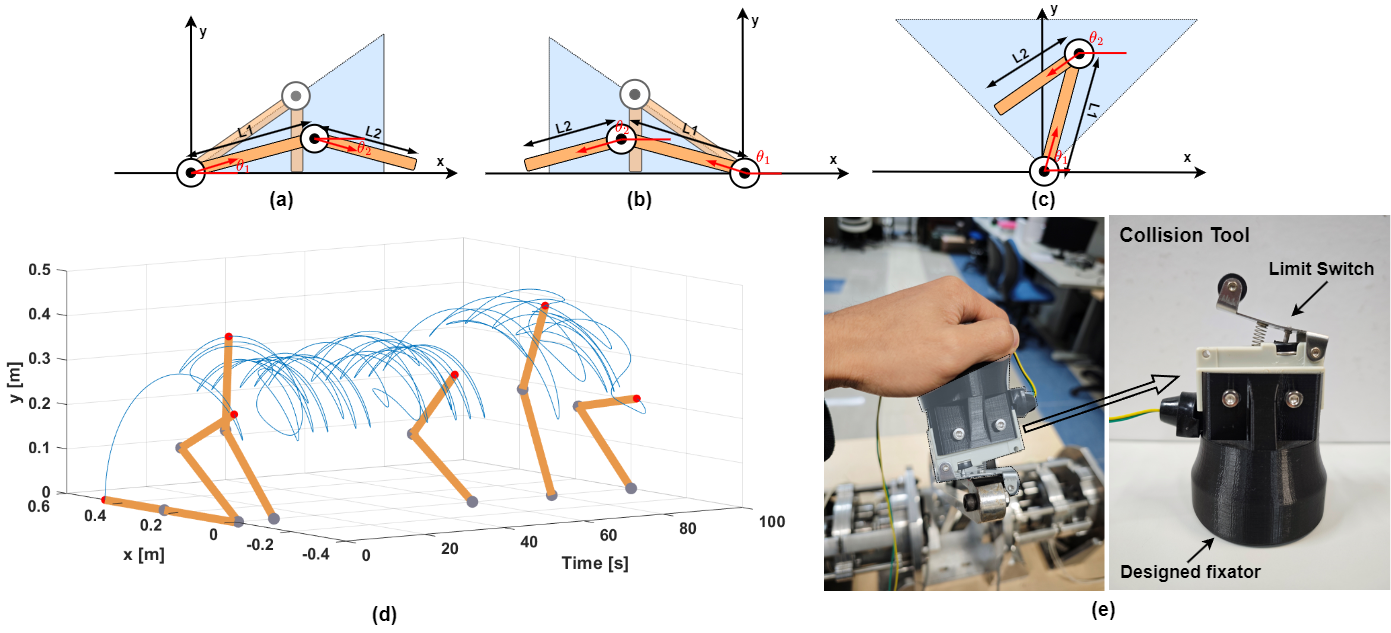}
    \caption{Three cases are considered regarding joint space constraints: (a) right-side constraints to prevent contact with the table, (b) left-side constraints for avoiding table contact, and (c) constraints aimed at preventing self-collision. (d) Random point-to-point acceleration-deceleration movements without table contacts and self-collision. (e) The designed collision tool, which mainly incorporates the limit switch, is used to collide with the manipulator and record the ground truth collision labels. It outputs binary results: 1 - collision, 0 - no collision.}
    \label{fig:joint space limitation} 
\end{figure}

The data collection process utilized the platform described in Section \ref{sec:platform introduction} for conducting experiments. During this process, the robot performed random point-to-point acceleration-deceleration movements within its joint space without implementing any collision reaction strategy. To achieve random point-to-point acceleration-deceleration movements without table contact and self-collision, constraints are established within the joint space. Three primary cases are analyzed concerning joint space constraints, as depicted in Figure \ref{fig:joint space limitation} (a)-(c). For added safety during manipulator movements, a margin $\theta_{m}$ is defined. This ensures the manipulator's secure operation by preventing collisions with both the table and self-collision. The overall constrained joint space $\theta$ is defined as:

% \begin{equation}
%     \theta_{2} \in \left\{
%     \begin{aligned}
%         & [\arcsin(l_{1}\sin(\theta_{1})/l_{2}) + \theta_{m}, \pi+\theta_1-\theta_m], \theta_1 \in [\theta_m, \arcsin(l_2/l_1)], \\
%         & [\theta_m-(\pi-\theta_1), \pi+\arcsin(l_1\sin(\pi-\theta_1)/l_2)], \theta_1 \in [\pi-\arcsin(l_2/l_1),\pi-\theta_m], \\
%         & [\theta_m-(\pi-\theta_1), \pi+\theta_1-\theta_m], \theta_1 \in [\arcsin(l_2/l_1), \pi-\arcsin(l_2/l_1)].
%     \end{aligned}
%     \right
% \end{equation}
\begin{equation}
    \theta_{2} \in \left\{
    \begin{aligned}
        & [\arcsin(l_{1}\sin(\theta_{1})/l_{2}) + \theta_{m}, \pi+\theta_1-\theta_m], && \theta_1 \in [\theta_m, \arcsin(l_2/l_1)], \\
        & [\theta_m-(\pi-\theta_1), \pi+\arcsin(l_1\sin(\pi-\theta_1)/l_2)], && \theta_1 \in [\pi-\arcsin(l_2/l_1),\pi-\theta_m], \\
        & [\theta_m-(\pi-\theta_1), \pi+\theta_1-\theta_m], && \theta_1 \in [\arcsin(l_2/l_1), \pi-\arcsin(l_2/l_1)].
    \end{aligned}
    \right.
\end{equation}

We employ a systematic sampling strategy within the Cartesian space, guided by the predefined constraints in the joint space. To facilitate more efficient collision data collection, we enable the robot to oscillate between the negative and positive x-axis. This is achieved by sampling end-effector target points in these two regions alternately. This approach not only ensures a comprehensive exploration of the constrained joint space, but also facilitates the systematic acquisition of collision data. Following the sampling of end-effector target points, we implement cubic spline data interpolation to enhance the smoothness of motion. To ensure precise position tracking, we employ a Proportional-Integral-Derivative (PID) controller. Additionally, gain scheduling is employed to facilitate accurate position tracking across varying stiffness conditions. This comprehensive approach results in the realization of random point-to-point acceleration-deceleration movements as shown in Figure \ref{fig:joint space limitation}(d).

During the collision data collection, a human operator deliberately collided with the robot at various random positions and instances using a collision tool (specifically, a limit switch) in order to record the ground truth data, where a value of 1 represents a collision, and 0 indicates no collision. The collision tool used in this data collection process is depicted in Figure \ref{fig:joint space limitation}(e). The video showing the process of collecting collision data is available in the Supplementary Materials. The joint space dataset (Joint torque $\tau$, joint velocity $\dot{\theta}$) was recorded at a sampling rate of 1000 Hz. The data collection process considers two distinct scenarios: collision and collision-free. Within each scenario, three cases are examined, representing different levels of physical stiffness (specifically, stiffness levels 2, 3, and 4). It is worth noting that the lowest stiffness level (stiffness level 1, as indicated in Figure \ref{fig:manipulator and DVSA working principle} (c)) is exclusively assigned to the safety configuration, following our previous work \cite{niu2022towards}. In particular, when a collision is detected, the stiffness of all joints is rapidly adjusted to the lowest level. Consequently, the data collection process does not involve the scenario associated with the lowest-level stiffness.

\begin{table}[b]
\centering
\caption{Training and testing dataset and the corresponding duration}
\label{Table:training and testing dataset time}
\begin{tabular}{|l|l|l|l|}
\hline
\rowcolor[HTML]{9B9B9B} 
Data Type                 & States                         & Joints Stiffness & Time  \\ \hline
Training                  & Collision                      & Highest level    & 4min  \\ \hline
                          &                                & Highest level    & 10min \\ \cline{3-4} 
                          &                                & 3rd level        & 10min \\ \cline{3-4} 
                          & \multirow{-3}{*}{Collision}    & 2nd level        & 10min \\ \cline{2-4} 
                          &                                & Highest level    & 15min \\ \cline{3-4} 
                          &                                & 3rd level        & 15min \\ \cline{3-4} 
\multirow{-6}{*}{Testing} & \multirow{-3}{*}{No Collision} & 2nd level        & 15min \\ \hline
\end{tabular}
\end{table}

Table \ref{Table:training and testing dataset time} presents the training and testing data scenarios, together with their respective durations. For training purposes, we only use collision data lasting four minutes, specifically from the highest stiffness level (level 4, as indicated in Figure \ref{fig:manipulator and DVSA working principle} (c)). The remaining data collected from different stiffness levels are used to evaluate the network's robustness to stiffness change. The testing phase involves a total of 30 minutes of collision motions (consisting of 516 collisions) and 45 minutes of collision-free motions.

%%%%%%%%%%%%%%%%%%%%%%%%%%%%%%%%%%%%%%%%%%%%%%%%%%%%%%%%%%%%%%%%%%%%%%%%%%%%%%%%%%%%%%%%%
% \subsection{Input Processing}
\begin{figure}[t]
    \centering
    \includegraphics[width=0.5\linewidth]{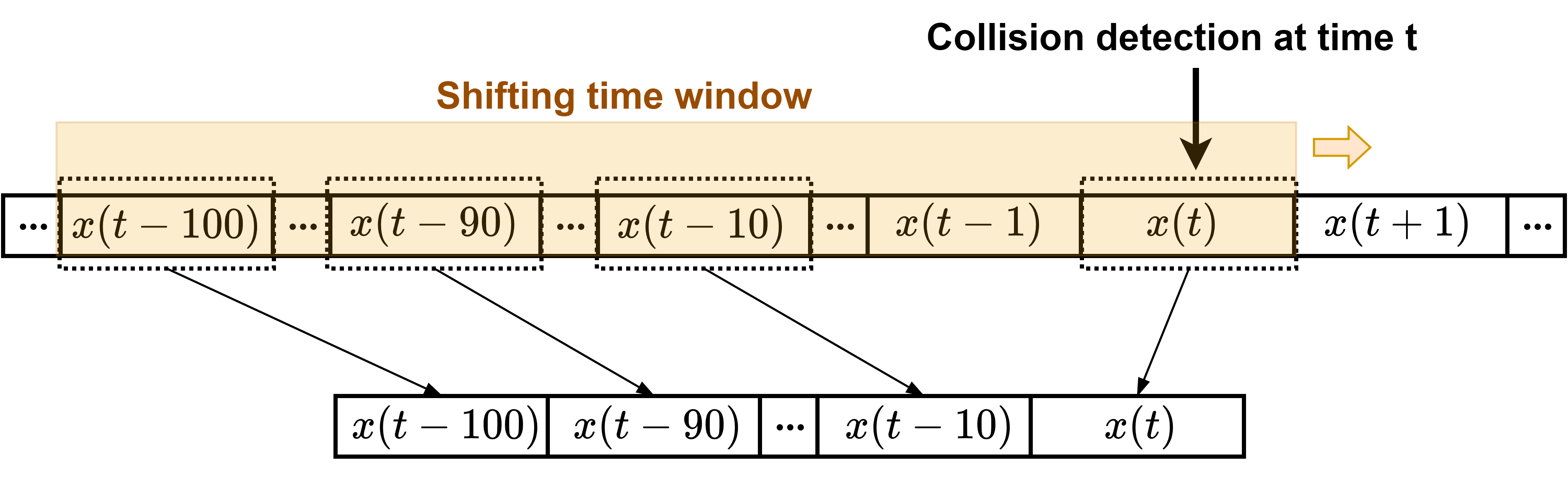}
    \caption{The time-series signals $\bold{x}(t)$ are segmented by dividing them into consecutive non-overlapping windows of size 101, with a sampling interval of 10. These segmented windows of $\bold{x}(t)$ are then sampled and arranged as the input data fed into the network.}
    \label{fig:time window}
\end{figure}

All the collected data is normalized to fit within the range of [0, 1]. Furthermore, as shown in Figure \ref{fig:time window}, the time-series data is segmented using a shifting time window that includes the current data point and the ten past points, with a sampling interval of $t_{I} = 0.01$. This results in a time window duration of $t_{w} = (11-1) \times 0.01 = 0.1s$. The utilization of segmented time-series data with a time window aids in mitigating the influence of sensor noise, as suggested by Kim \textit{et al}. \cite{kim2021transferable}. The choice of time step eleven is determined through an evaluation process where the time step is progressively increased until a relatively superior performance is achieved. Then, the input to MAD-CNN is shaped as:
\begin{equation}
    X = [X^1, X^2],
\end{equation}
$X^{i}$ represents the input data corresponding to the $i$th joint. These $X^{i}$ data are constructed by combining the segmented data, as depicted in Figure \ref{fig:time window} and explained as
\begin{equation}
    X^{i} = [x^{i}(t-100), x^{i}(t-90),..., x^{i}(t-10), x^{i}(t)],
\end{equation}
and $x^{i}(t)$ includes the normalized observed signals, which consist of the joint torque $\tau$ and joint velocity $\dot{\theta}$:
\begin{equation}
    x^{i}(t) = [\tau, \dot{\theta}].
\end{equation}

%%%%%%%%%%%%%%%%%%%%%%%%%%%%%%%%%%%%%%%%%%%%%%%%%%%%%%%%%%%%%%%%%%%%%%%%%%%%%%%%%%%%%%%%%
% \subsection{Network Training Details}
Training data in Table \ref{Table:training and testing dataset time} are randomly shuffled prior to training. The network structure is implemented in Python with PyTorch 1.13.1. The mini-batch size is 1000, and the number of epochs is set to 30. The optimizer is Adam with default parameters ($\beta_{1}$ = 0.9, $\beta_{2}$ = 0.999, $\epsilon$ = 1e-07) and the learning rate of $lr$ = 1e-3. We use binary cross-entropy loss (BCELoss) as the loss function. 

\begin{equation}
    \text{BCELoss} = -(y * log(\hat{y}) + (1 - y) * log(1 - \hat{y})),
\end{equation}
where $y$ represents the ground truth collision label, while $\hat{y}$ denotes the network's prediction.

%%%%%%%%%%%%%%%%%%%%%%%%%%%%%%%%%%%%%%%%%%%%%%%%%%%%%%%%%%%%%%%%%%%%%%%%%%%%%%%%%%%%%%%%%
\section{Results}
\label{sec: results}
\begin{figure}[t]
    \centering
    \includegraphics[width=0.45\linewidth]{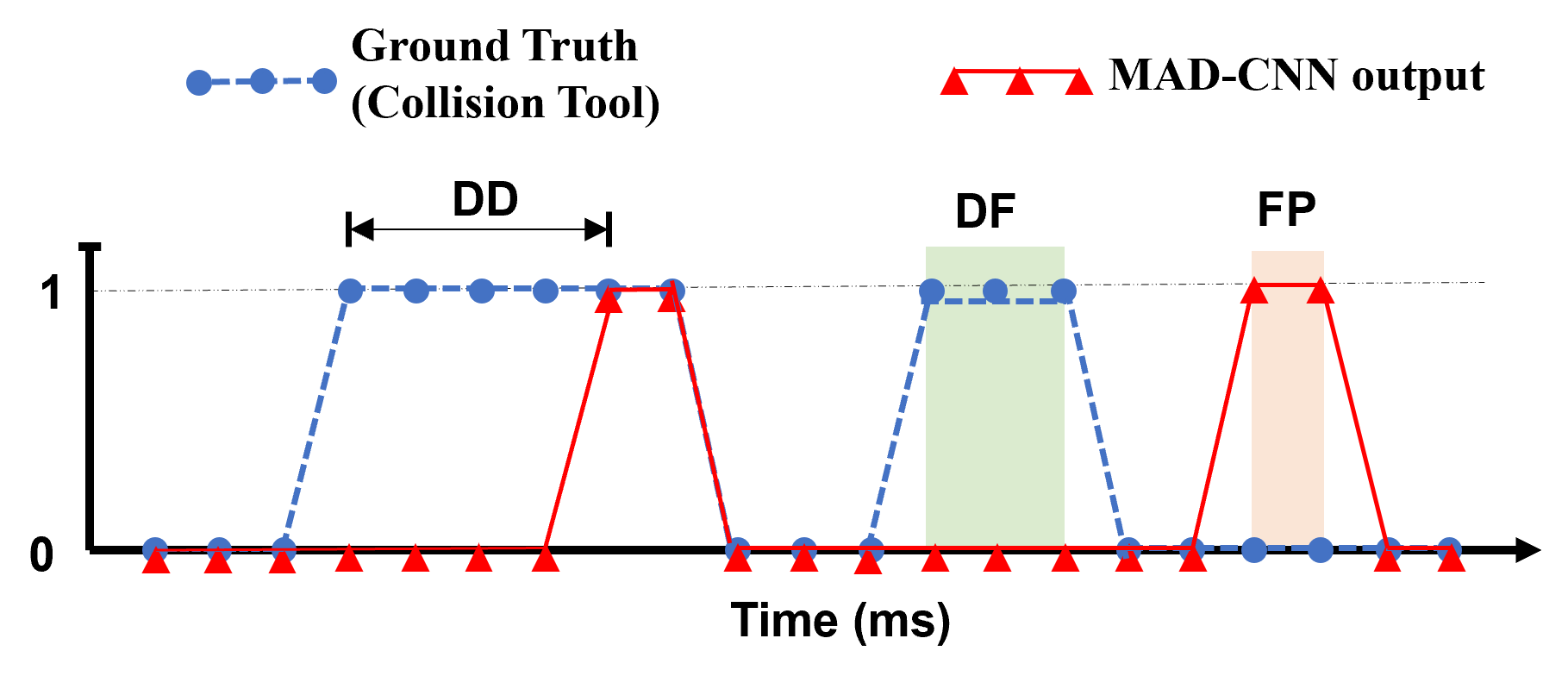}
    \caption{Explanation of Detection Delay (DD), Detection Failure (DF) and False Positive (FP). The dots represent the samples}
    \label{fig:DD_DF_FP_explanation}
\end{figure}

To evaluate the proposed approach and conduct a fair comparison with state-of-the-art methods, we use three criteria shown in Figure \ref{fig:DD_DF_FP_explanation}: detection failure number (DFn), detection delay (DD) in milliseconds (ms), and false positive number (FPn), consistent with the evaluation framework in \cite{kim2021transferable}. The collision tool generates a "1" signal when a collision occurs, and the network predicts collisions with an output threshold of 0.5. DFn represents the count of collisions not accurately classified by the network, while FPn indicates instances where the network wrongly classifies non-collision events as collisions. Minimizing FPn is crucial for operational efficiency, as false positives may trigger unnecessary stop actions. Successive FPs within the time step are considered as one FP.

To address FP, a continuous filter (CF) is commonly used, as highlighted in \cite{kim2021transferable, park2020learning}. While CF can reduce FP, it may increase DFn and DD, creating a trade-off between collision detection sensitivity and false positive occurrence. Section \ref{sec:FP handling with CF} investigates the impact of integrating CF into MAD-CNN to assess its efficiency in enhancing collision detection robustness. An ablation study in Section \ref{sec:ablation study} evaluates the contributions of various components within MAD-CNN, verifying the superiority of the proposed network architecture. Additionally, in Section \ref{sec:network comparison}, we perform a comparative analysis of collision detection performance between MAD-CNN and state-of-the-art models.

%%%%%%%%%%%%%%%%%%%%%%%%%%%%%%%%%%%%%%%%%%%%%%%%%%%%%%%%%%%%%%%%%%%%%%%%%%%%%%%%%%%%%%%%%%%%

\subsection{Ablation Study}
\label{sec:ablation study}
To assess the individual contributions of key components within MAD-CNN, we conducted an ablation study. Five network configurations were evaluated. 

\textbf{1. MAD-CNN (full model)}: Includes all components -- modularization, dilated convolution, and attention module. 

\textbf{2. M-CNN (modularization only)}: Uses modularization but excludes dilated convolution and attention. 

\textbf{3. MD-CNN (modularization and dilation only)}: Excludes the attention module while retaining modularization and dilated convolution. 

\textbf{4. MA-CNN (modularization and attention only)}: Omits dilated convolution but retains modularization and attention. 

\textbf{5. AD-CNN (joint input)}: Eliminates modularization and uses all joint signals as one input to the network.

Table \ref{Table: Ablation study} summarizes the performance of each configuration, demonstrating the individual impact of each component on collision detection accuracy, detection delay, and false positive rate.
We proceed to examine the individual contributions of these components in greater detail.

\paragraph{Modularization} This approach effectively reduces the parameter search space and enhances data efficiency during training. As evident in the comparison between MAD-CNN and AD-CNN, modularization contributes to mitigating detection failures and delays. However, M-CNN, which relies solely on modularization, struggles to achieve satisfactory collision sensitivity.
\begin{table}[t]
\centering
\caption{Results for ablation study of MAD-CNN}
\label{Table: Ablation study}
\begin{tabular}{|cll|l|l|l|l|l|}
\hline
\rowcolor[HTML]{C0C0C0} 
\multicolumn{3}{|l|}{\cellcolor[HTML]{C0C0C0}Ablation Study}                                                                                                             & \textbf{MAD-CNN} & M-CNN        & MD-CNN  & MA-CNN  & AD-CNN  \\ \hline
\multicolumn{3}{|l|}{Modularization}                                                                                                                                     & \checkmark                & \checkmark            & \checkmark       & \checkmark       &         \\ \hline
\multicolumn{3}{|l|}{Dilation}                                                                                                                                           & \checkmark                &              & \checkmark       &         & \checkmark       \\ \hline
\multicolumn{3}{|l|}{Attention}                                                                                                                                          & \checkmark                &              &         & \checkmark       & \checkmark       \\ \hline
\rowcolor[HTML]{FFFFFF} 
\multicolumn{1}{|c|}{\cellcolor[HTML]{FFFFFF}}                                   & \multicolumn{1}{l|}{\cellcolor[HTML]{FFFFFF}}                                & DFn    & \textbf{0/172}   & 0/172        & 0/172   & 4/172   & 0/172   \\ \cline{3-8} 
\rowcolor[HTML]{FFFFFF} 
\multicolumn{1}{|c|}{\cellcolor[HTML]{FFFFFF}}                                   & \multicolumn{1}{l|}{\cellcolor[HTML]{FFFFFF}}                                & DD(ms) & \textbf{11.3095} & 15.3491      & 15.1065 & 18.1818 & 13.6488 \\ \cline{3-8} 
\rowcolor[HTML]{FFFFFF} 
\multicolumn{1}{|c|}{\cellcolor[HTML]{FFFFFF}}                                   & \multicolumn{1}{l|}{\multirow{-3}{*}{\cellcolor[HTML]{FFFFFF}Highest level}} & FPn    & 183              & \textbf{42}  & 74      & 543     & 158     \\ \cline{2-8} 
\rowcolor[HTML]{FFFFFF} 
\multicolumn{1}{|c|}{\cellcolor[HTML]{FFFFFF}}                                   & \multicolumn{1}{l|}{\cellcolor[HTML]{FFFFFF}}                                & DFn    & \textbf{0/172}   & 2/172        & 1/172   & 10/172  & 0/172   \\ \cline{3-8} 
\rowcolor[HTML]{FFFFFF} 
\multicolumn{1}{|c|}{\cellcolor[HTML]{FFFFFF}}                                   & \multicolumn{1}{l|}{\cellcolor[HTML]{FFFFFF}}                                & DD(ms) & \textbf{12.4093} & 17.6         & 16.3508 & 19.7592 & 14.4011 \\ \cline{3-8} 
\rowcolor[HTML]{FFFFFF} 
\multicolumn{1}{|c|}{\cellcolor[HTML]{FFFFFF}}                                   & \multicolumn{1}{l|}{\multirow{-3}{*}{\cellcolor[HTML]{FFFFFF}3rd level}}     & FPn    & 261              & \textbf{102} & 116     & 135     & 167     \\ \cline{2-8} 
\rowcolor[HTML]{FFFFFF} 
\multicolumn{1}{|c|}{\cellcolor[HTML]{FFFFFF}}                                   & \multicolumn{1}{l|}{\cellcolor[HTML]{FFFFFF}}                                & DFn    & \textbf{0/172}   & 24/172       & 1/172   & 30/172  & 6/172   \\ \cline{3-8} 
\rowcolor[HTML]{FFFFFF} 
\multicolumn{1}{|c|}{\cellcolor[HTML]{FFFFFF}}                                   & \multicolumn{1}{l|}{\cellcolor[HTML]{FFFFFF}}                                & DD(ms) & \textbf{12.4319} & 18.6283      & 16.7894 & 21.4647 & 16.0304 \\ \cline{3-8} 
\rowcolor[HTML]{FFFFFF} 
\multicolumn{1}{|c|}{\multirow{-9}{*}{\cellcolor[HTML]{FFFFFF}Joints Stiffness}} & \multicolumn{1}{l|}{\multirow{-3}{*}{\cellcolor[HTML]{FFFFFF}2nd level}}     & FPn    & 137              & \textbf{78}  & 94      & 120     & 123     \\ \hline
\multicolumn{2}{|c|}{}                                                                                                                                          & DFn    & \textbf{0/516}   & 26/516       & 2/516   & 44/516  & 6/516   \\ \cline{3-8} 
\multicolumn{2}{|c|}{}                                                                                                                                          & DD(ms) & \textbf{12.0502} & 17.1924      & 16.0822 & 19.8019 & 14.6934 \\ \cline{3-8} 
\multicolumn{2}{|c|}{\multirow{-3}{*}{Total}}                                                                                                                   & FPn    & 581              & \textbf{222} & 284     & 798     & 448     \\ \hline
\end{tabular}
\end{table}

\paragraph{Dilated Convolution} This technique strengthens collision detection by extracting discriminative features with long-range dependencies. Compared to MA-CNN, MAD-CNN's improved sensitivity and robustness can be attributed to dilation. However, it might slightly increase false positives at lower stiffness levels due to enhanced sensitivity to system noise. Similar observations are present when comparing M-CNN and MD-CNN.

\paragraph{Attention Module} Prioritizing relevant features through the attention mechanism significantly improves collision detection, reducing both the detection failure number and the delay (DFn and DD). However, it may introduce a small increase in false positives due to sensitivity to system noise. In particular, combining modularization and attention without dilation, as in the MA-CNN case, demonstrates suboptimal performance compared to other configurations.

The ablation study reveals that each component of MAD-CNN plays a crucial role in its superior performance. Modularization fosters data efficiency and reduces detection delays, while dilated convolution enhances sensitivity and robustness. The attention module further improves collision detection, but slightly increases false positives. Overall, the synergistic combination of these components in MAD-CNN yields the best results in terms of accuracy, speed, and robustness for collision detection across diverse stiffness conditions.

%%%%%%%%%%%%%%%%%%%%%%%%%%%%%%%%%%%%%%%%%%%%%%%%%%%%%%%%%%%%%%%%%%%%%%%%%%%%%%%%%%%%%%%%%%%%
\subsection{False Positive Handling with Continuous Filter}
\label{sec:FP handling with CF}

While MAD-CNN outperformed other networks in terms of detection failure and delay (Section \ref{sec:ablation study}), it exhibited a slightly higher false positive (FP) rate. This section investigates the effectiveness of a continuous filter in addressing false positives while minimizing impact on detection accuracy and speed. The CF, a temporal filtering mechanism, assigns the collision label only when the network prediction exceeds a specific duration threshold. While reducing FPs, CF can increase detection delay and detection failure number. For instance, \cite{kim2021transferable} reported a 20. 5\% increase in DD with a 3ms CF in their MNN model.

To optimize CF use for MAD-CNN, 
\begin{figure}[t]
    \centering
    \includegraphics[width=0.9\linewidth]{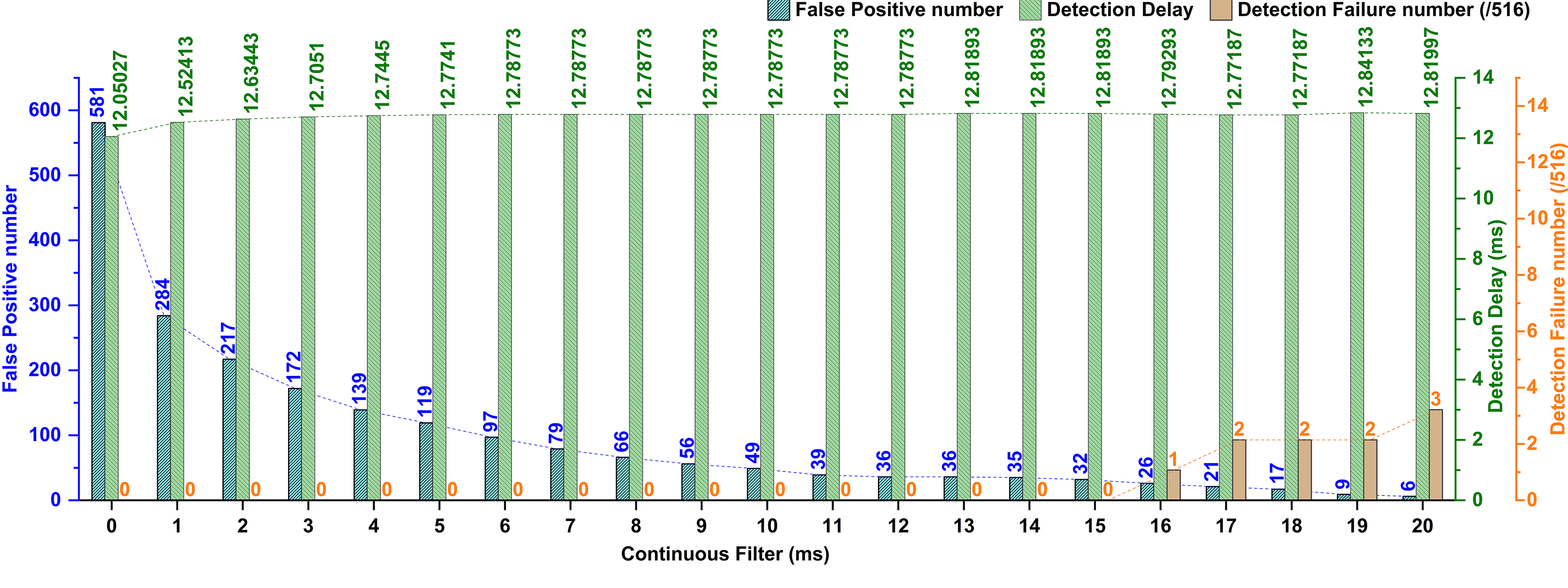}
    \caption{The study of continuous filter (CF) integrated with MAD-CNN. We explore various CF durations ranging from 0ms to 20ms to assess the robustness of MAD-CNN for collision detection.}
    \label{fig:CF effects}
\end{figure}
we evaluated various durations from 0 to 20ms across all stiffness levels. Figure \ref{fig:CF effects} summarizes the results. 

\textbf{False Positives:} As expected, FPn decreased significantly with increasing duration of CF. Notably, only 6 FPs were observed with a 20ms CF, representing a substantial reduction compared to the unfiltered case. 

\textbf{Detection Delay:} Interestingly, DD did not increase dramatically with longer CF durations. The maximum increase (6.5\%) was observed with a 19ms CF, and even at 20ms, DD remained stable at around 12.8ms. 

\textbf{Detection Failure:} MAD-CNN maintained exceptional detection accuracy (100\%) with CF durations up to 15ms. Beyond this, DFn gradually increased, but even with a 20ms CF, the model achieved 99.4\% accuracy (3 failures out of 512 collisions).

These findings demonstrate MAD-CNN's robustness and sensitivity to collisions even when using a CF. Importantly, a 15ms CF offers a beneficial trade-off between FP reduction and minimal impact on DD and DFn. Therefore, we adopt this duration for further analysis in the next section.

%%%%%%%%%%%%%%%%%%%%%%%%%%%%%%%%%%%%%%%%%%%%%%%%%%%%%%%%%%%%%%%%%%%%%%%%%%%%%%%%%%%%%%%%%%%%

\subsection{Performance Comparison with Existing Methods}
\label{sec:network comparison}
\begin{figure}[t]
    \centering
    \includegraphics[width=0.9\linewidth]{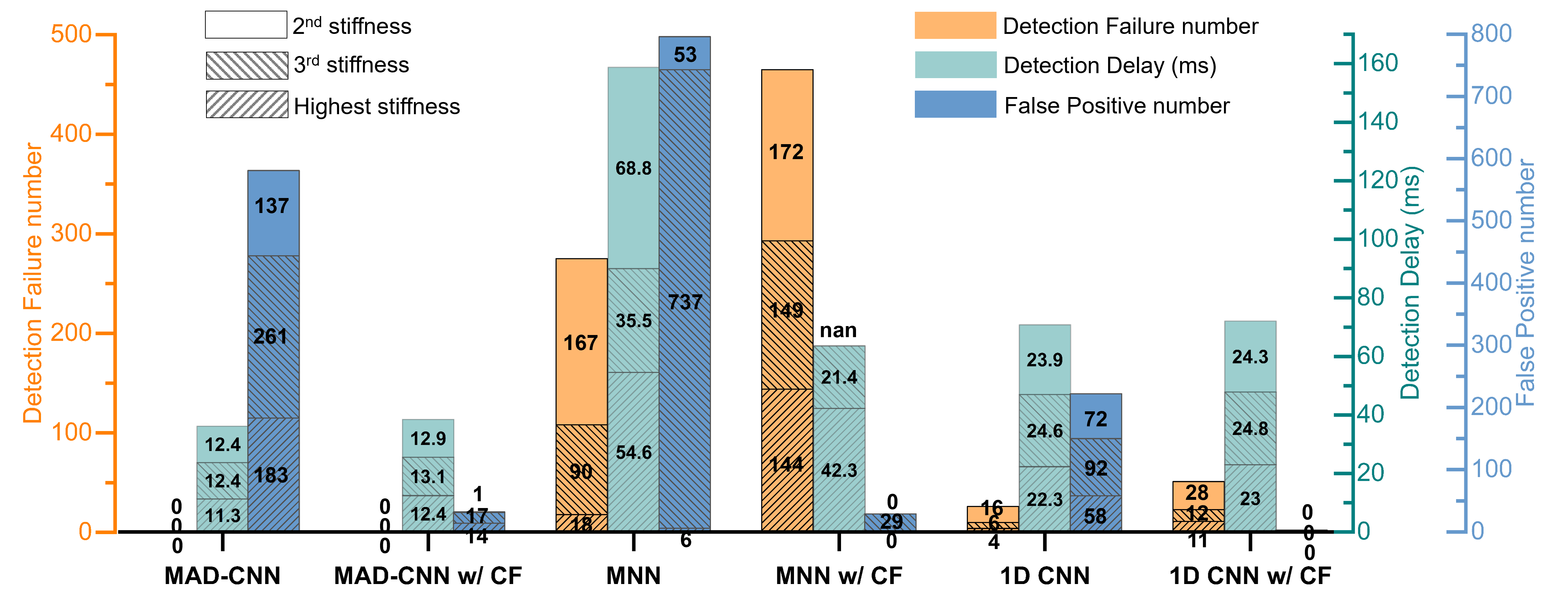}
    \caption{Performance comparison is conducted against existing methods, namely MAD-CNN, MNN, and 1D-CNN. Furthermore, an evaluation is made for these three methods incorporating continuous filters (w/ CF), considering metrics such as detection failure count (172 collisions per stiffness, totaling 516 collisions), detection delay in milliseconds, and false positive count.}
    \label{fig:comparison with other methods}
\end{figure}

This section benchmarks MAD-CNN against two prominent learning-based collision detection methods: MNN \cite{kim2021transferable}, employed in Doosan robot M0609, and the 1D CNN model \cite{park2020learning}. To comprehensively evaluate sensitivity, robustness, and false positive reduction, we utilize a 15ms continuous filter in the comparison (Figure \ref{fig:comparison with other methods}).

MAD-CNN stands out, achieving \textbf{100\% collision detection} with an average delay of \textbf{12.0503ms}. The 1D CNN demonstrates good sensitivity, while MNN falls behind, likely due to its limited training data. Although CF effectively reduces false positives across all methods, it slightly impacts detection failure number and detection delay, particularly for 1D CNN. Notably, MAD-CNN maintains outstanding robustness with only a \textbf{6.4\% increase in DD} and a significant reduction in false positives. Furthermore, despite training on just 4 minutes of data from the highest stiffness setting, MAD-CNN exhibits remarkable robustness to stiffness variations. It maintains \textbf{100\% accuracy} across all stiffness levels, with only minor delay increases under different conditions. This superior performance across both data efficiency and adaptability underlines MAD-CNN's promising potential for real-world deployment in variable stiffness robots.

\begin{figure}[t]
    \centering
    \includegraphics[width=\linewidth]{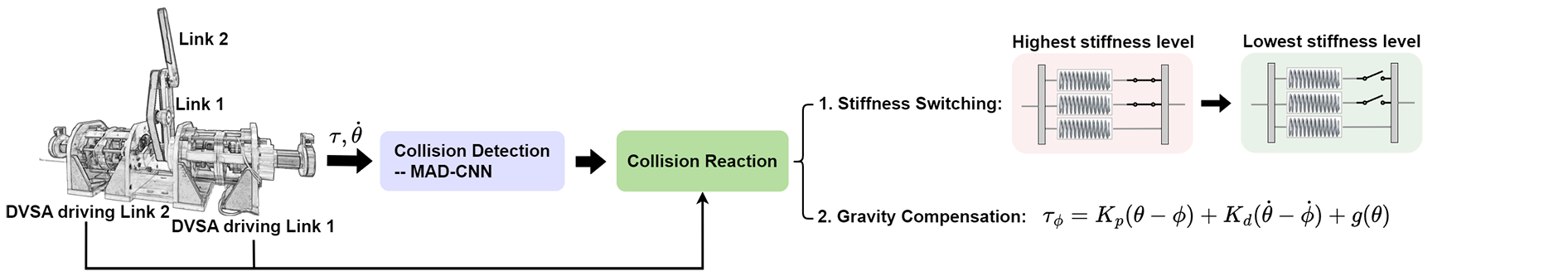}
    \caption{The post-collision handling pipeline with the integration of the proposed MAD-CNN and DVSA-based manipulator to effectively handle unexpected collisions with data efficiency and robustness.}
    \label{fig:post-collision handling pipeline}
\end{figure}

In the pursuit of ensuring safety during physical human-robot interaction, the integration of rapid and precise collision detection and response mechanisms is essential. Building upon our prior research \cite{niu2022towards}, the fusion of collision detection with a DVSA-based manipulator reaction has proven to be effective in mitigating unforeseen collisions, thereby enhancing safety. The proposed MAD-CNN demonstrates the ability to achieve accurate and robust collision detection performance with data efficiency, eliminating the need for a robot model. The integration of MAD-CNN contributes to the enhancement of the post-collision handling pipeline's performance. The framework is depicted in Figure \ref{fig:post-collision handling pipeline}. Upon MAD-CNN detecting a collision, two simultaneous reactions are initiated. The stiffness of the two actuators rapidly changes to the lowest, simultaneously switching the tracking mode to gravity compensation with proportional-derivative (PD) control in the joint space, as expressed below:
\begin{equation}
    \tau_{\phi} = K_{p}(\theta-\phi)+K_{d}(\dot{\theta}-\dot{\phi})+g(\theta),
\end{equation}
where $\theta$ and $\phi$ represent the position vectors of the link side and motor side, respectively. $\tau_{\phi}$ denotes the vector of motor input torques, with $K_{p}$ and $K_{d}$ as positive symmetric definite matrices. The term $g(\theta)$ represents the vector of gravity effects. Within this framework, a rapid, more accurate, and robust post-collision handling pipeline is integrated to enhance safety during HRI.

While MAD-CNN demonstrates impressive performance in variable stiffness robots, exploring its potential and robustness further requires investigation in two key directions:

\begin{itemize}
    \item \textbf{Expanding Evaluation Scope:} 
    Evaluating MAD-CNN in commercial robots with multiple rigid joints constitutes a valuable next step. Such an assessment would unveil its generalizability across diverse robotic architectures, beyond variable stiffness, and reveal potential adaptations for broader applicability.
    
    \item \textbf{Exploring Unsupervised Learning:}
    Leveraging unsupervised learning techniques holds promise for reducing reliance on labeled collision data. Autoencoders (AEs) offer a potential path for data-efficient training, enabling them to extract intrinsic features without explicit labels. Additionally, generative adversarial networks (GANs) can generate synthetic collision scenarios, augmenting training datasets and mitigating data limitations. Exploring these methods could lead to more robust and adaptable collision detection models applicable to a wider range of robotic platforms.

\end{itemize}

%%%%%%%%%%%%%%%%%%%%%%%%%%%%%%%%%%%%%%%%%%%%%%%%%%%%%%%%%%%%%%%%%%%%%%%%%%%%%%%%%%%%%%%%%%%%
\section{CONCLUSIONS}
\label{sec:conclusion}

This paper introduces MAD-CNN, a novel collision detection network for robots with variable stiffness actuators. By leveraging a dual inductive bias mechanism and an attention module, MAD-CNN achieves exceptional performance with minimal data requirements. Notably, it trains efficiently on just 4 minutes of collision data, even from a single stiffness setting. MAD-CNN exhibits remarkable robustness to stiffness variations, as evidenced by extensive experiments showcasing its accurate collision detection capabilities across diverse stiffness conditions. This resilience holds even when trained on limited data from a singular stiffness setting, rendering MAD-CNN an ideal solution for robots with variable stiffness actuators, particularly in scenarios featuring limited and imbalanced collision data.
%MAD-CNN demonstrates remarkable robustness to stiffness variations. Extensive experiments showcase its ability to accurately detect collisions across diverse stiffness conditions, even when trained on limited data from a single stiffness setting. This resilience makes MAD-CNN an ideal solution for robots equipped with variable stiffness actuators, particularly in scenarios with limited and imbalanced collision data.
The comparative analysis further highlights MAD-CNN's superiority over existing learning-based models. It surpasses state-of-the-art methods in terms of collision detection sensitivity, robustness, and data efficiency. These combined strengths position MAD-CNN as a promising candidate for real-world implementation in variable stiffness robots, fostering advanced capabilities and enhanced safety in diverse applications.

\section{Acknowledgements}

This work was supported by Khalifa University of Science and Technology under Award No. RC1-2018-KUCARS and FSU-2021-019.

\section{Author contributions}

Z.N.: Conceptualization, Data curation, Investigation, Methodology, Formal analysis, Writing-original draft and editing. L.S.S.: Conceptualization, Data curation, Investigation, Methodology, Formal analysis, Writing-review and editing. I.H.: Supervision, Funding acquisition, Writing-review and editing.

\section{Data availability}
Data will be available on request. Please contact Zhenwei Niu (zhenwei.niu@ku.ac.ae) to request this data.
%\section{Competing interests}
%The author(s) declare no competing interests.

% \section*{REFERENCES}
\bibliographystyle{chicago}
\bibliography{references}

\end{document}